\documentclass[runningheads]{llncs}
\usepackage[T1]{fontenc}
\usepackage{graphicx}
\usepackage{tikz}
\usetikzlibrary{shapes.geometric, arrows, positioning, arrows.meta, calc, backgrounds, matrix}
\usepackage{float}
\usepackage{colortbl}
\usepackage{diagbox}
\usepackage{tcolorbox}
\usepackage{xcolor}
\usepackage{enumitem}
\usepackage{amsmath}
\usepackage{amssymb}
\usepackage{rotating}
\usepackage{cite}

\definecolor{notsolightgreen}{RGB}{200,255,200}
\definecolor{green(html/cssgreen)}{rgb}{0.0, 0.5, 0.0}

\definecolor{lightgreen}{RGB}{230,255,230}
\definecolor{lightred}{RGB}{255,230,230}
\definecolor{lightyellow}{RGB}{255,255,230}
\definecolor{darkgreen}{RGB}{0,150,0}
\definecolor{darkred}{RGB}{150,0,0}
\definecolor{deepyellow}{RGB}{180,140,0}

\newcommand{\goodmark}{\cellcolor{lightgreen}\textcolor{darkgreen}{\checkmark}}
\newcommand{\badmark}{\cellcolor{lightred}\textcolor{darkred}{\textsf{X}}}
\newcommand{\maybemarkbg}{\cellcolor{lightyellow}\textcolor{deepyellow}{(\checkmark)}}

\newtcolorbox{searchbox}[1]{
	colback=lightgreen,
	colframe=darkgray,
	fonttitle=\bfseries,
	title=#1,
	before skip=10pt,
	after skip=10pt
}
\begin{document}
\title{Overview of AI Grading of Physics Olympiad Exams}
%\author{First Author\inst{1}\orcidID{0000-1111-2222-3333}} 
%\institute{X \and Y
\author{Lachlan McGinness\inst{1,2}\orcidID{0000-0002-3231-4827}} 
\authorrunning{L. McGinness}
\institute{Australian National University \and
CSIRO, Australia\\
\email{lachlan.mcginness@anu.edu.au}\\
}

\maketitle              % typeset the header of the contribution
\begin{abstract}
Automatically grading the diverse range of question types in high school physics problem is a challenge that requires automated grading techniques from different fields. We report the findings of a Systematic Literature Review of potential physics grading techniques. We propose a multi-modal AI grading framework to address these challenges and examine our framework in light of Australia's AI Ethical Principles. 
\end{abstract}
\section{Introduction}

Studies show that teachers in Australia are burning out due to their workloads \cite{Windle2022Teachers, AITSL2023Australia}. At both the institutional and individual level, automated marking Artificial Intelligence (AI) solutions are starting to be sought out and employed in schools to reduce this workload \cite{Ogg2024Brisbane}. %The specific details of automated marking solutions are rarely disclosed by companies to avoid criticism and loss of “trade secrets”. 
Studying potential AI Marking solutions from both an ethical and performance perspective allows governments, schools, teachers and students to make informed decisions about which of these tools can ethically and safely be used to reduce teacher workload.

In this paper we explore the automated grading of high school level physics problems, focusing on the Australian Physics Olympiad. Physics problems typically take a diverse range question types which could require any combination of these to mark. Typical physics question types are:
\begin{itemize}
    \item \textbf{Numerical} - Student response is a number with or without units and/or direction. For example $7$, $15$ m/s or $20$ m upwards.
    \item \textbf{Algebraic} - A question which requires students to give their answer in the form of an algebraic expression, equation or proof. For example $\frac{L_0}{4 \pi r^2}$ or $x+y = x^2 + y^2$.
    \item \textbf{Plots or Graphs} - Student response is a plot of data or function.
    \item \textbf{Diagrams} - Student response is a visual representation that does not fall into plots or graph.
    \item \textbf{Short Answer} - A worded response, normally to a closed-ended question, evaluated on technical correctness or use of specific key terms. For example ``The friction force is acting in the upward direction''.
    \item \textbf{Multiple Choice} - A question where students select one or more provided options. These are very simple to grade and not considered in this study.
\end{itemize}

\noindent We propose a Multi-Modal AI grading framework which uses a range of techniques to grade these diverse problem types. This paper reports on a December 2024 Systematic Literature Review (SLR) of automated marking to explore the range of tools that could be used to grade physics questions. We make two contributions; we summarise key findings from the SLR and we position potential grading solutions within Australia's Ethical AI Principles. 

\section{Systematic Literature Review}

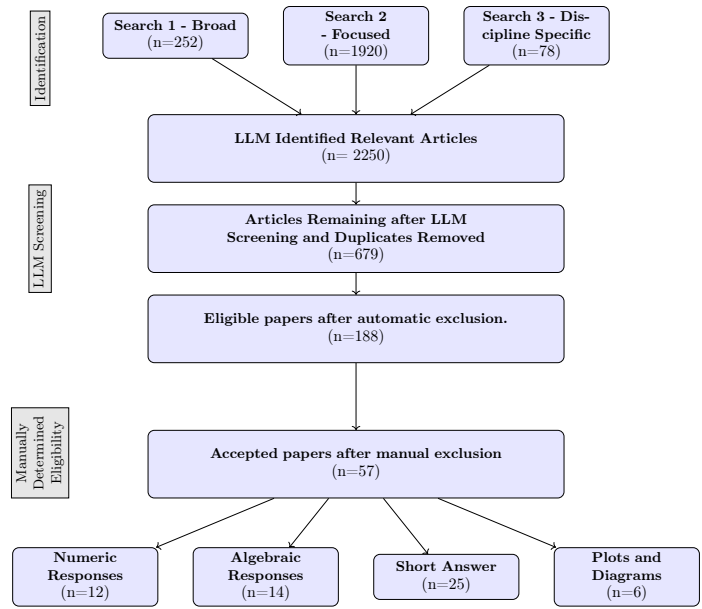
\begin{figure}[htb]
	\centering
    \scalebox{0.6}{
	\begin{tikzpicture}[scale=1, transform shape,
		auto,
		box/.style={
			draw,
			rectangle,
			rounded corners,
			fill=blue!10,  % Light blue fill
			minimum width=3cm,
			minimum height=1cm,
			text width=3cm,
			align=center
		},
		widebox/.style={
			draw,
			rectangle,
			rounded corners,
			fill=blue!10,  % Light blue fill
			minimum width=7cm,
			minimum height=1.5cm,
			text width=9cm,
			align=center
		},
		phase/.style={
			draw,
			rectangle,
			minimum width=2cm,
			fill=gray!20,
			rotate=90,
			align=center
		}
		]
		% Define the starting y coordinate
		\def\starty{0}
		
		% Phase labels on the left
		\node[phase] at (-7,-0.5) {Identification};
		\node[phase] at (-7,-4.5) {LLM Screening};
		\node[phase] at (-7,-9.25) {Manually\\ Determined\\ Eligibility};
		
		% Top row - three search boxes, more spread out
		\node[box] (search1) at (-4,\starty) {\footnotesize \textbf{Search 1 - Broad}\\ \normalsize (n=252)};
		\node[box] (search2) at (0,\starty) {\footnotesize \textbf{Search 2 - Focused}\\ \normalsize (n=1920)};
		\node[box] (search3) at (4,\starty) {\footnotesize \textbf{Search 3 - Discipline Specific}\\ \normalsize (n=78)};
		
		% Second level - identified articles (wider)
		\node[widebox] (identified) at (0,\starty-2.5) 
		{\textbf{LLM Identified Relevant Articles}\\ \normalsize(n= 2250)};
		
		% Arrows from search boxes to identified
		\draw[->] (search1) -- (identified);
		\draw[->] (search2) -- (identified);
		\draw[->] (search3) -- (identified);
		
		% Screening level (wider)
		\node[widebox] (screening) at (0,\starty-4.5) 
		{\textbf{Articles Remaining after LLM Screening and Duplicates Removed}\\ \normalsize (n=679)};
		
		\draw[->] (identified) -- (screening);
		
		\node[widebox] (AutomaticallyExcluded) at (0,\starty-6.5) 
		{\textbf{Eligible papers after automatic exclusion.}\\ \normalsize (n=188)};
		
		\draw[->] (screening) -- (AutomaticallyExcluded);
		
		\node[widebox] (ManuallyExcluded) at (0,\starty-9.5) 
		{\textbf{Accepted papers after manual exclusion}\\ \normalsize (n=57)};
		
		\draw[->] (AutomaticallyExcluded) -- (ManuallyExcluded);
		
		\node[box] (numeric) at (-6,\starty-12) {\textbf{Numeric Responses}\\ \normalsize (n=12)};
		\node[box] (algebra) at (-2,\starty-12) {\textbf{Algebraic Responses}\\ \normalsize (n=14)};
		\node[box] (shortAnswer) at (2,\starty-12) {\textbf{Short Answer}\\ \normalsize (n=25)};
		\node[box] (plots) at (6,\starty-12) {\textbf{Plots and Diagrams}\\ \normalsize (n=6)};
		
		\draw[->] (ManuallyExcluded) -- (numeric);
		\draw[->] (ManuallyExcluded) -- (algebra);
		\draw[->] (ManuallyExcluded) -- (shortAnswer);
		\draw[->] (ManuallyExcluded) -- (plots);
		
	\end{tikzpicture}
    }
	\caption{Flow diagram illustrating the steps taken in the literature review and the number of papers included in each step. Details of the search terms and databases for each of the three sources can be found in Appendix \ref{appendix:Search Terms}.}
	\label{fig:prisma}
\end{figure}

In December 2024, a systematic literature review was conducted using the steps as illustrated in Figure \ref{fig:prisma}. Databases were searched for relevant papers. Then papers were automatically screened to a focused set, building on techniques described in \cite{McGinness2024Highlighting}. Finally the papers were manually screened to ensure only those which were peer reviewed and would contribute to physics marking were considered.

\noindent The automatic screening process used a Large Language Model to determine the type of questions that each paper addressed as well as the methods used. The upcoming subsections give the key findings of the literature review for each question type commonly used in physics exams. A further breakdown of the statistics of the papers examined can be found in Appendix \ref{appendix:SummaryStatistics}.

\subsection{Numeric Questions}
When students type their inputs in a fixed format, numerical questions are easy to grade using rule based approaches. The technique is now easily implemented in forms and spreadsheets \cite{McNeil2015Excel}.

\iffalse

\begin{table}[H]
\centering
    \caption{Rule based approaches deal with approximate and exact answers, units and directions for strictly formatted inputs. %Marking handwritten responses can be computationally expensive and does not allow complex mathematical expressions. Reliably extracting student answers, including intermediate steps from free-form text is still an open problem.
    }
    \label{tab:NumericalMarkingProblems}
	\scalebox{0.8}{\centering
		\setlength{\tabcolsep}{4pt}
		\setlength{\arrayrulewidth}{1pt}
		\begin{tabular}{|p{4cm}|p{3cm}|p{3cm}|p{3cm}|}
			\hline
			\multicolumn{4}{|c|}{\textbf{\normalsize Outline of Problems in Numerical Questions}} \\
			\hline
			\small \textbf{Numerical Question Challenge} & 
			\small \textbf{Solved Problem} & 
			\small \textbf{Addressed in Literature} & 
			\small \textbf{Required in Physics Marking} \\
			\hline
			\small Exact Answers & \goodmark & \goodmark & \goodmark \\
			\hline
			\small Rounding/Approximate & \goodmark & \goodmark & \goodmark \\
			\hline
			\small Units & \maybemarkbg & \goodmark & \goodmark \\
			\hline
			\small Directions (Vectors) & \maybemarkbg & \goodmark & \goodmark \\
			\hline
			\small Intermediate Steps & \badmark & \maybemarkbg & \goodmark \\
			\hline
			\small Extracting Final Answer from freeform response & \badmark & \badmark & \goodmark \\
			\hline
			\small Hand Written Responses & \maybemarkbg & \goodmark & \goodmark \\
			\hline
	\end{tabular}}
\end{table}
\vspace{-0.5cm}
\fi

Handwritten responses require Optical Character Recognition (OCR) techniques. In 2015, Xerox developed a system which automatically marked hand-written numeric responses \cite{Gross2015Handwriting}. In 2024 a new dataset MNIST-Fraction was released as a benchmark for measuring accuracy of recognising fractions \cite{Ahadian2024MNIST}.

A more recent approach focused on using Large Multi-modal Models (LMMs) to grade student responses to calculus questions with numerical answers \cite{Gandolfi2024GPT}. The best performing available model (GPT4) often made numerical mistakes and the logic behind its grading was not strong, though the grades correlated with human markers. Feature extraction and explainable Machine Learning (ML) methods have been used to grade numerical responses in physics \cite{McGinness2024CONFOLD}. 

%As displayed in Table \ref{tab:NumericalMarkingProblems}, 
Numerical responses are more difficult to mark if units or directions (for vector quantities) are included, as commonly required in physics. Answers can be provided in particular units and a pre-defined coordinate system to mitigate this problem. However if free form answers are accepted, then feature extraction techniques as discussed in Sections \ref{sec:algebra} and \ref{sec:shortResponse} will be required.

\subsection{Algebraic Questions}

\label{sec:algebra}
Marking algebraic expressions, equations, or proofs tends to be a more difficult task than marking numerical responses as there are many different correct forms of the answer. Consider a question as simple solving Kepler's third law for period:
\begin{equation*}
	T = \sqrt{\frac{4 \pi^2 r^3}{GM}} \text{ or } T = \sqrt{\frac{4 \pi^2 r}{GM}}r \text{ or } T = \sqrt{\frac{4 \pi^2}{GMr}}r^2  \text{ or } T = 2\pi \sqrt {\frac{r^3}{GM}} \text{ or ...} %\text{ or } T = 2 \sqrt{\frac{ \pi^2 r^3}{GM}}
\end{equation*}

\noindent When expressions are more complex, determining equivalence can be arduous. When they include trigonometric expressions and exponentiation it is impossible to find a general algorithm that verifies the equivalence of two expressions \cite{Richardson1968Some}. 

Computer Algebra systems such as SymPy \cite{Sympy} implement heuristics that will often succeed in determining equivalence, but they are rigid and require students to understand the system's syntax. Therefore examiners are measuring the students' mastery of the the system's syntax as well as the course content \cite{Fast2013Crowd}. 
%This may be the reason why few papers have been published implementing Computer Algebra systems in education, (see Figures \ref{fig:methods-heatmap} and \ref{fig:techniques-histogram} in Appendix \ref{appendix:SummaryStatistics}).

Rule-based parsing systems that require examiners to anticipate student errors do not scale well \cite{McNichols2023Algebra}. Large Language Models (LLMs) can be used to capture and translate meaningful features from student responses into a computer readable format \cite{McNichols2023Algebra}. Kortemeyer used LLMs for grading direclty \cite{Kortemeyer2023Toward}, but found that despite correlation between LLM and human graders, LLMs miss nuance. 

\subsection{Plots, Graphs and Diagrams}

There are few studies which grade hand drawn diagrams or plots and graphs. Li et al. deployed a small, open-source, multi-modal, visual-reasoning LLM to provide coherent feedback on images in mathematical settings \cite{Li2024Automated}. Kortemeyer et al. used LMMs to mark physics exams involving diagrams and found that in 2024, LLMs struggled to distinguish essential details from irrelevant information \cite{Kortemeyer2024Grading}. They also found that LMMs made more errors when the number of rubric items increased, consistent with studies in adjacent domains \cite{McGinness2024Highlighting}. 

\subsection{Short Answer}
\label{sec:shortResponse}

Many Automated Short Answer Grading (ASAG) techniques follow two steps:
\begin{enumerate}
    \item A feature extraction technique, such as Bag of Words or Part of Speech Tagging, converts the text into a mathematical representation. 
    \item Then a ML algorithm such as cosine similarity or Term Frequency Inverse Document Frequency is applied to this representation. 
\end{enumerate}
LLMs have also been applied to automated grading of short answers \cite{Kortemeyer2024Automated}, but the literature has found they have limitations. For example, knowing the human assigned mark causes Large Language Models to bias their own marks \cite{Grevisse2024LLM-based}.

\section{AI Ethics}
\vspace{-0.25cm}
Most automated grading studies do not explicitly consider ethics. We turn to the Australian AI Ethics Principles \cite{AustraliaAIEthics} to evaluate the ethical dimensions. Three of the relevant principles for are Human and Environmental Well-being, Privacy, and Contestability. In order to meet these, AI grading systems should:
\begin{itemize}
    \item take flexible inputs and require no training to reduce teacher workload.
    \item minimise environmental impact and respect privacy by running locally.
    \item  use explainable, not-secret methods with guarantees (be reliable).
\end{itemize}  
Table \ref{tab:SLRmethodsSummary} summarises four methods with the question types they can address.
\vspace{-0.5cm}
\begin{table}[htbp]
	\caption{Application of ethical principles to automated grading techniques.}
	\label{tab:SLRmethodsSummary}
	\centering
	\setlength{\tabcolsep}{1pt}
	\setlength{\arrayrulewidth}{1pt} 
    \scalebox{0.8}{
	\begin{tabular}{|p{3.2cm}|*{13}{c|}}
		\hline
		\diagbox[
		width=3cm,
		height=8mm,
		linewidth=1pt,
		innerwidth=3cm,
		innerleftsep=0pt,
		innerrightsep=5pt
		]{Technique}{Criteria}  & 
		\multicolumn{1}{p{7mm}|}{\begin{turn}{90}\scriptsize Reliable\end{turn}} & 
		\multicolumn{1}{p{7mm}|}{\begin{turn}{90}\scriptsize Explainable\end{turn}} & 
		\multicolumn{1}{p{7mm}|}{\begin{turn}{90}\scriptsize No Training Data\end{turn}} & 
		\multicolumn{1}{p{7mm}|}{\begin{turn}{90}\scriptsize No Expert Time\end{turn}} & 
		\multicolumn{1}{p{7mm}|}{\begin{turn}{90}\scriptsize Open Source\end{turn}} & 
		\multicolumn{1}{p{7mm}|}{\begin{turn}{90}\scriptsize Maintains Privacy\end{turn}} & 
		%\multicolumn{1}{p{7mm}|}{\begin{turn}{90}\scriptsize Able to Run on Laptop\end{turn}} & 
		\multicolumn{1}{p{7mm}|}{\begin{turn}{90}\scriptsize Runs Locally\end{turn}} & 
		\multicolumn{1}{p{7mm}|}{\begin{turn}{90}\scriptsize Flexible inputs\end{turn}} & 
		\multicolumn{1}{p{7mm}|}{\begin{turn}{90}\scriptsize Numeric \end{turn}} & 
		\multicolumn{1}{p{7mm}|}{\begin{turn}{90}\scriptsize Algebraic \end{turn}} & 
		\multicolumn{1}{p{7mm}|}{\begin{turn}{90}\scriptsize Short Answer\end{turn}} & 
		\multicolumn{1}{p{7mm}|}{\begin{turn}{90}\scriptsize Plots and Graphs\end{turn}} & 
		\multicolumn{1}{p{7mm}|}{\begin{turn}{90}\scriptsize Diagrams\end{turn}}  \\
		%\multicolumn{1}{p{7mm}|}{\begin{turn}{90}\scriptsize Criterion 12\end{turn}} & 
		%\multicolumn{1}{p{7mm}|}{\begin{turn}{90}\scriptsize Criterion 13\end{turn}} & 
		%\multicolumn{1}{p{7mm}|}{\begin{turn}{90}\scriptsize Criterion 14\end{turn}} & 
		%\multicolumn{1}{p{7mm}|}{\begin{turn}{90}\scriptsize Criterion 15\end{turn}} \\
		\hline
		\footnotesize{Rule based methods} & \goodmark & \goodmark & \goodmark & \badmark & \goodmark &  \goodmark &  \goodmark & \badmark & \goodmark & \maybemarkbg & \maybemarkbg &  \badmark & \badmark \\
		\hline
		\footnotesize{Hand-crafted features} & \goodmark & \goodmark & \goodmark & \badmark & \maybemarkbg & \goodmark & \goodmark & \badmark & \goodmark & \goodmark & \goodmark & \badmark &  \badmark \\
		\hline
		\footnotesize{Feature Extraction and Machine Learning}  & \badmark & \maybemarkbg & \badmark & \goodmark & \goodmark & \goodmark & \goodmark & \goodmark & \goodmark & \goodmark & \goodmark & \badmark &  \badmark \\
		\hline
		%\footnotesize{Automated Feature Extraction and Neural Networks} & \goodmark & \badmark & \badmark & \goodmark & \goodmark & \goodmark & \maybemarkbg & \goodmark & \goodmark & \goodmark & \goodmark & \badmark & \badmark  \\
		%\hline
		\footnotesize{Multi-modal Large Language Models} & \badmark & \maybemarkbg & \goodmark & \goodmark & \maybemarkbg  & \maybemarkbg & \maybemarkbg & \goodmark & \goodmark & \maybemarkbg & \maybemarkbg & \maybemarkbg &   \maybemarkbg \\
		\hline
		%\footnotesize{Traditional Computer Vision} & \goodmark & \badmark & \badmark & \maybemarkbg & \goodmark & \goodmark & \maybemarkbg & \badmark & \maybemarkbg & \maybemarkbg & \maybemarkbg & \maybemarkbg &  \maybemarkbg \\
		%\hline
		%Method 7 & & \goodmark & \badmark & \maybemarkbg & \goodmark & \badmark & \maybemarkbg & \goodmark & \badmark & \maybemarkbg & \goodmark & \badmark & \maybemarkbg  \\
		\hline
	\end{tabular}}
\end{table}
\normalsize 

\section{Current Challenges, Limitations and Future Work}
\vspace{-0.25cm}
In order to meet the ethical criteria across the range of question types, further work will explore deployment of an LLM-modulo framework \cite{Subbarao2024LLMmodulo}. This uses verification techniques to improve LLM reliability, including those which use external tools \cite{McGinness2024Automated}. 

This literature review covers a very broad field of question types and marking methods, but more narrow research still needs to be conducted to be aware of the most promising techniques compatible with an LLM Modulo approach. 
AIED, ITS, Learning at Scale and LAK conference proceedings were not included in this review, additional work should be done to find relevant techniques from these venues.

Despite some initial studies, the accuracy of LLMs in screening papers is still largely unknown \cite{McGinness2024Highlighting}. A significant number of the papers could be checked manually to determine if the error rate when screening is sufficiently accurate for Systematical Literature Review. 

In order to be confident that no relevant papers were excluded a significant number of the papers could be checked manually to attempt to determine the error rate and determine if this is a sufficiently accurate process for systematical literarture review. 

Once reliable pre-processing is applied, Computer Algebra Systems can be used to verify students' expressions and proofs. Australian Physics Olympiad problems typically require algebraic manipulation involving some transcendental functions, but not calculus. An area of future work is to identify which reasoning systems are most accurate and efficient for responses of this form. 

\begin{credits}
\subsubsection{\ackname} This project has been funded by the CSIRO and ANU.
\vspace{-0.25cm}
\subsubsection{\discintname}
The authors have no competing interests to declare. % that are relevant to the content of this article. 
\end{credits}
\vspace{-0.25cm}
%\section*{Bilbiography}

\bibliographystyle{splncs04}
\bibliography{PhDThesisReferences}

\appendix

\break

\section{Details of Search Terms}
\label{appendix:Search Terms}

\begin{figure}[H]
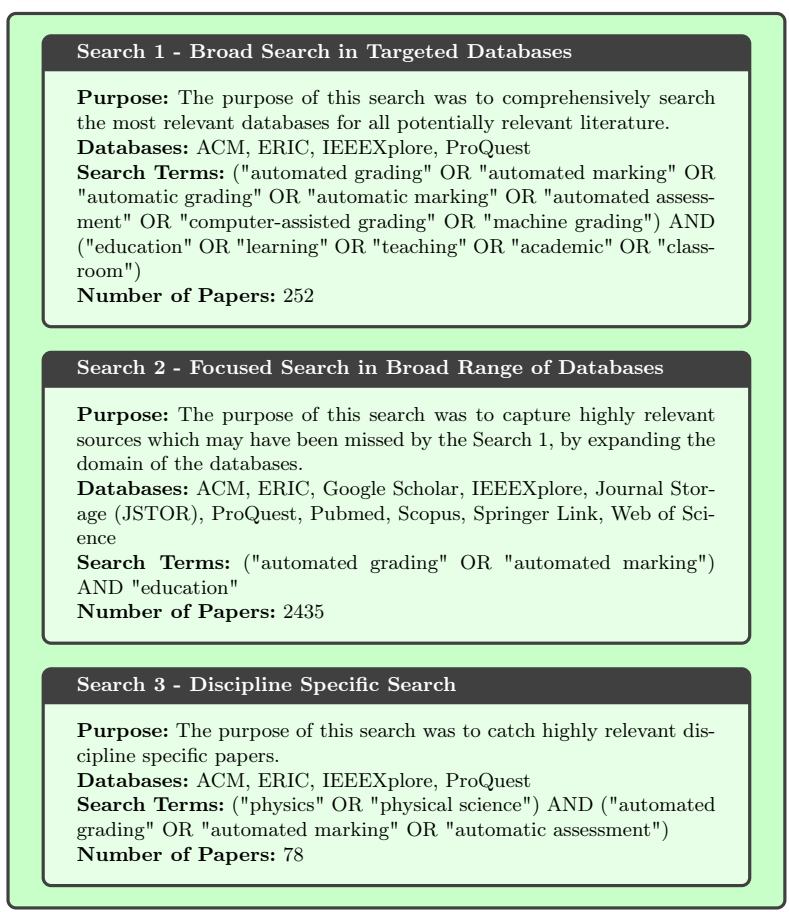

	\centering
	\scalebox{0.85}{
		\begin{tcolorbox}[
			colback=notsolightgreen,
			%sharp corners,
			colframe=darkgray
			]
			\footnotesize
			\begin{searchbox}{Search 1 - Broad Search in Targeted Databases}
				\textbf{Purpose:} The purpose of this search was to comprehensively search the most relevant databases for all potentially relevant literature. 
				
				\textbf{Databases:} ACM, ERIC, IEEEXplore, ProQuest
				
				\textbf{Search Terms:} ("automated grading" OR "automated marking" OR "automatic grading" OR "automatic marking" OR "automated assessment" OR "computer-assisted grading" OR "machine grading") AND ("education" OR "learning" OR "teaching" OR "academic" OR "classroom")
				
				\textbf{Number of Papers:} 252
			\end{searchbox}
			
			\begin{searchbox}{Search 2 - Focused Search in Broad Range of Databases}
				\textbf{Purpose:} The purpose of this search was to capture highly relevant sources which may have been missed by the Search 1, by expanding the domain of the databases.
				
				\textbf{Databases:} ACM, ERIC, Google Scholar, IEEEXplore, Journal Storage (JSTOR), ProQuest, Pubmed, Scopus, Springer Link, Web of Science
				
				\textbf{Search Terms:} ("automated grading" OR "automated marking") AND "education"
				
				\textbf{Number of Papers:} 2435
			\end{searchbox}
			
			\begin{searchbox}{Search 3 - Discipline Specific Search}
				\textbf{Purpose:} The purpose of this search was to catch highly relevant discipline specific papers.
				
				\textbf{Databases:} ACM, ERIC, IEEEXplore, ProQuest
				
				\textbf{Search Terms:} ("physics" OR "physical science") AND ("automated grading" OR "automated marking" OR "automatic assessment")
				
				\textbf{Number of Papers:} 78
			\end{searchbox}
			
		\end{tcolorbox}
	}
	\normalsize
	\caption{Details of search strategy used in the the Systematic Literature Review. Overall three searches were conducted, each with a different focus. Note that the databases include conference papers and books in addition to journal papers.}
	\label{fig:search-strategy}
\end{figure}

Three separate searches were conducted. The purpose of the primary search was to find all relevant articles from the most relevant databases. It used a broad set of search terms on four targeted databases which were most likely to be relevant for education literature: ACM Digital Library (ACM), Education Resources Information Centre (ERIC), IEEEXplore and ProQuest. The secondary search used a narrower set of keywords but searched across a broader base of datatbases (10 in total), trying to find the most relevant papers from these sources. The third search attempted to find discipline-specific papers that may been missed in the first search. More details about each of the searches can be found in Figure~\ref{fig:search-strategy}.

\pagebreak

\section{Statistics from Literature Review}
\label{appendix:SummaryStatistics}

\begin{figure}[H]
	\centering
	\includegraphics[width=\textwidth]{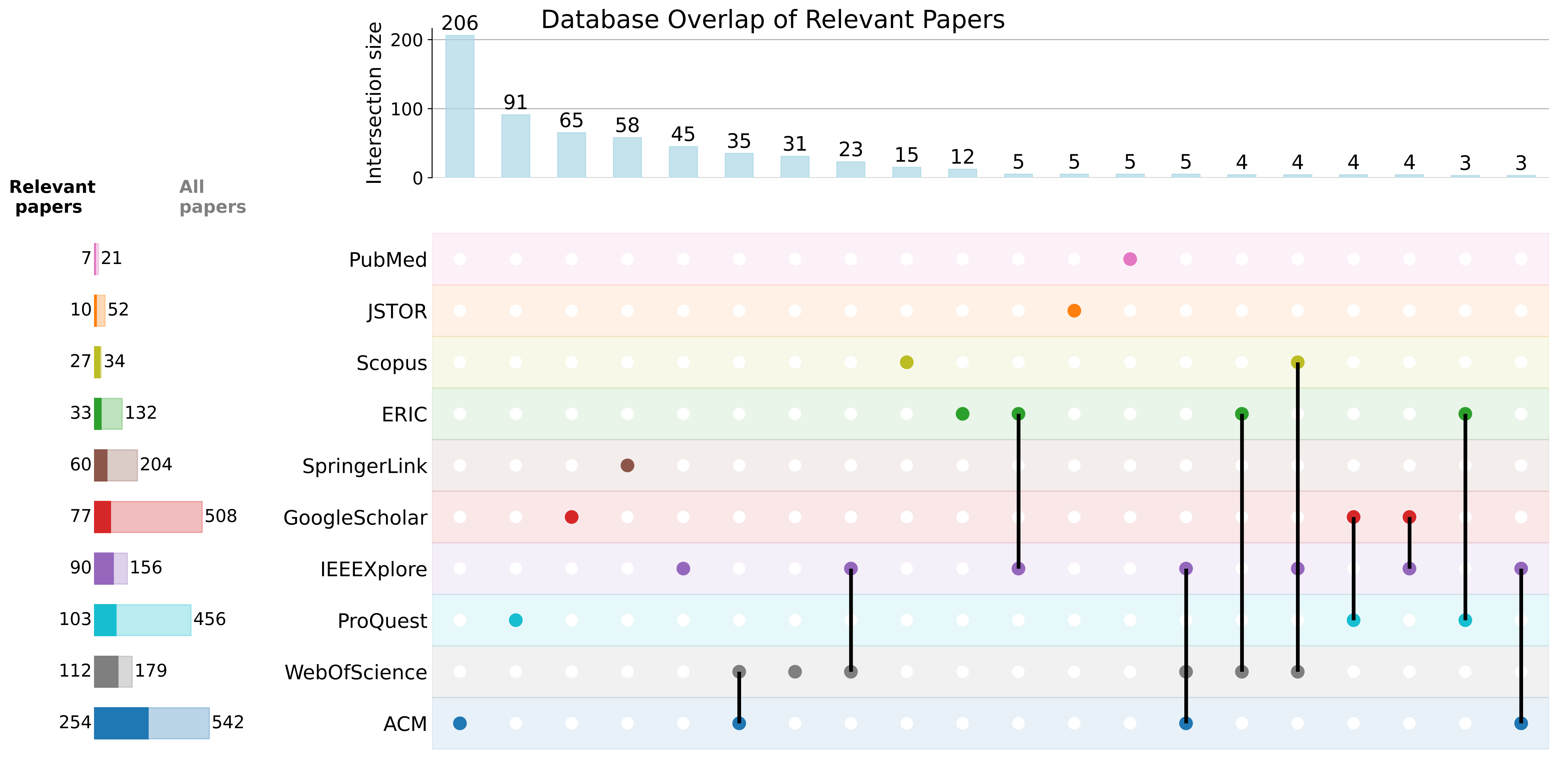}
	\caption{Overlap of research papers found across different academic databases. The horizontal bar chart (left) shows the total number of papers from each source. Dark coloured bars indicate relevant papers while lighter coloured bars show the total number of papers retrieved. The upset plot (right) shows the overlap between each database.}
	\label{fig:database-overlap}
\end{figure}

Although there were overlaps in the papers from each of the ten sources used in the three searches, each database had many papers that were not found in any of the others. This is illustrated in Figure \ref{fig:database-overlap}.  The three searches combined returned a large number of papers, bordering on the line of what is achievable for an individual to review manually. However, only some of these papers would be relevant for the study as shown in Figures \ref{fig:prisma} and \ref{fig:database-overlap}. Building on techniques from \cite{McGinness2024Highlighting}, LLMs were used to screen these papers, determine which were relevant and extract useful information. 

\begin{figure}[H]
	\centering
    \includegraphics[width=0.95\textwidth]{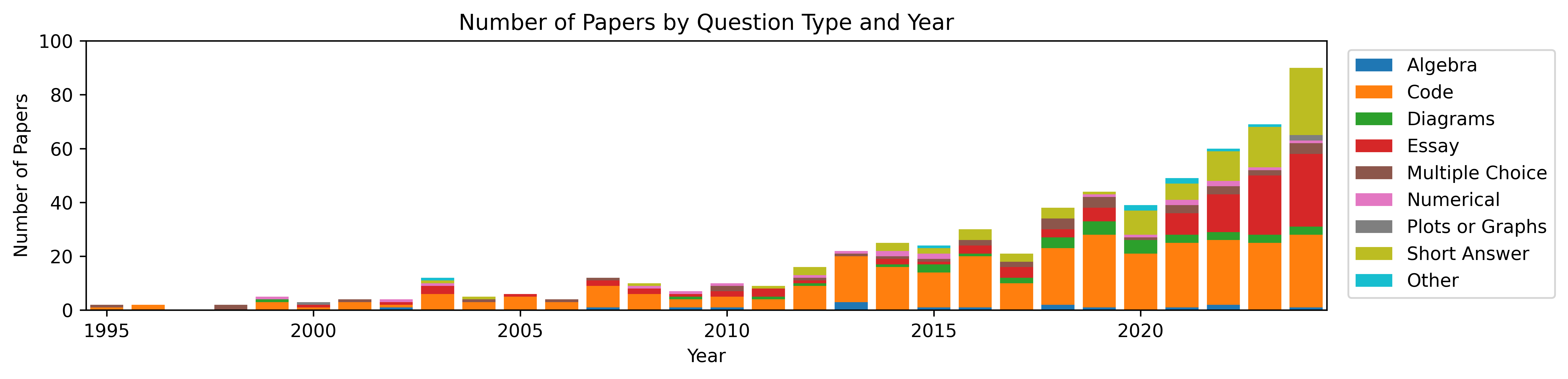}
    \includegraphics[width=0.95\textwidth]{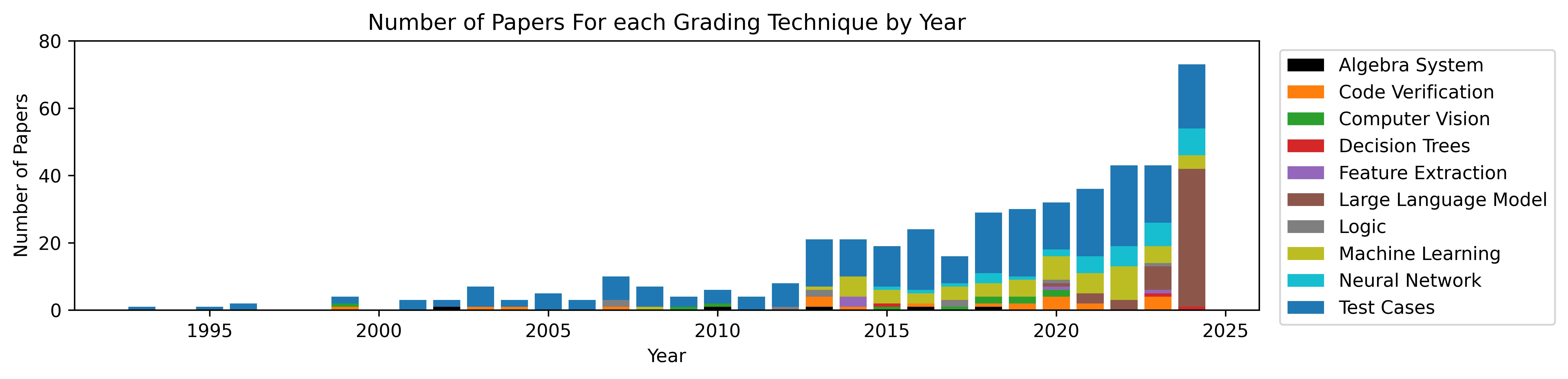}
	\caption{Histogram of the number of papers published for each automated grading technique and question type by year.}
	\label{fig:techniques-histogram}
\end{figure}

Figure \ref{fig:techniques-histogram} shows the overall trend in the types of questions and methods of automated grading addressed in papers over time. We can see a steady increase in the number of papers published each year. Historically, most papers discuss the automated grading of code-based questions but recently there has been an increase in essay and short answer questions.

Examining the number of papers for each grading technique by year of publication reveals an interesting trend over time, see Figure \ref{fig:techniques-histogram}. Test cases was by far the most common method used in automated grading dating back to the 1990s. Then in 2014, machine learning techniques were starting to be applied. In 2021 neural networks were starting to gain popularity before an explosion in Large Language Model automated grading in 2024. If this trend continues we should see that Large Language Model automated grading will be the dominant technique over from 2025 and beyond.

\vspace{-0.5cm}

\begin{figure}[htbp]
	\centering
	\includegraphics[width=0.9\textwidth]{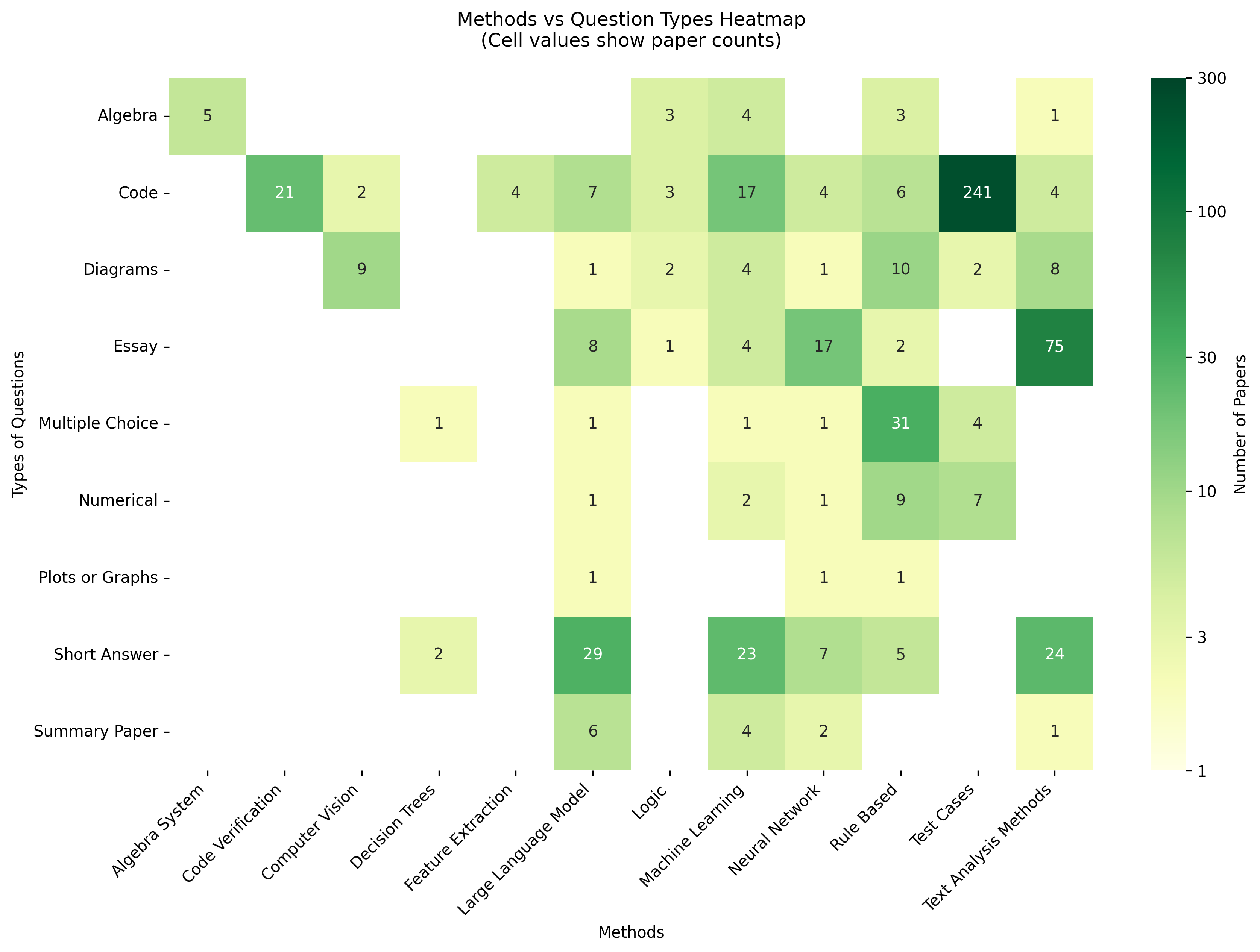}
	\caption{Heatmap showing the distribution of different methods used across various question types in automated assessment systems. The colour intensity represents the logarithmic count of papers, while the numbers in each cell show the actual count. Empty cells indicate no papers were found for that combination.}
	\label{fig:methods-heatmap}
\end{figure}

Figure \ref{fig:methods-heatmap} shows how commonly each method was used for each type of question. Unsurprisingly, text analysis methods were commonly used for essay questions and test cases were used for code. 

\vspace{-0.5cm}

\end{document}